\documentclass[10pt,twocolumn,letterpaper]{article}

\usepackage{iccv}
\usepackage{times}
\usepackage{epsfig}
\usepackage{bm}
\usepackage{graphicx}
\usepackage{amsmath}
\usepackage{amssymb}

\usepackage[breaklinks=true,bookmarks=false]{hyperref}
\usepackage{authblk}
\iccvfinalcopy 

\def\iccvPaperID{3820} 

\begin{document}

\title{End-to-End Learning Using Cycle Consistency \\ for Image-to-Caption Transformations}

\author[1]{Keisuke Hagiwara}
\author[1,2]{Yusuke Mukuta}
\author[1,2]{Tatsuya Harada}
\affil[1]{The University of Tokyo}
\affil[2]{RIKEN AIP}
\affil[ ]{\tt\small {\{hagiwara, mukuta, harada\}@mi.t.u-tokyo.ac.jp}}

\maketitle

\begin{abstract}
   So far, research to generate captions from images has been carried out from the viewpoint that a caption holds sufficient information for an image.
   If it is possible to generate an image that is close to the input image from a generated caption, i.e., if it is possible to generate a natural language caption containing sufficient information to reproduce the image, then the caption is considered to be faithful to the image. To make such regeneration possible, learning using the cycle-consistency loss is effective. In this study, we propose a method of generating captions by learning end-to-end mutual transformations between images and texts. To evaluate our method, we perform comparative experiments with and without the cycle consistency. The results are evaluated by an automatic evaluation and crowdsourcing, demonstrating that our proposed method is effective. 
\end{abstract}

\section{Introduction}

Generating captions from images is useful, and research has been conducted to solve this problem using machine learning. 
For example, there are many situations where it is difficult for blind people to correctly understand images on the web. During Internet shopping, it is necessary to obtain information on images of products in audio or text form to shop as expected. In addition, image captioning is required for non-blind users. For example, when the only information a user possesses concerning the content they are interested in is an image, it would be helpful for obtaining information via a search if it is possible to input this image and acquire text for the search. Moreover, when it is necessary to utilize words only to describe a captured scene, it is necessary to convey the situation captured by the camera in text form. In addition, applications for providing assistance based on the road condition while driving a car or understanding the situation in a room can also be considered. 

\begin{figure}[t]
 \centering
 \includegraphics[width=.4\textwidth]{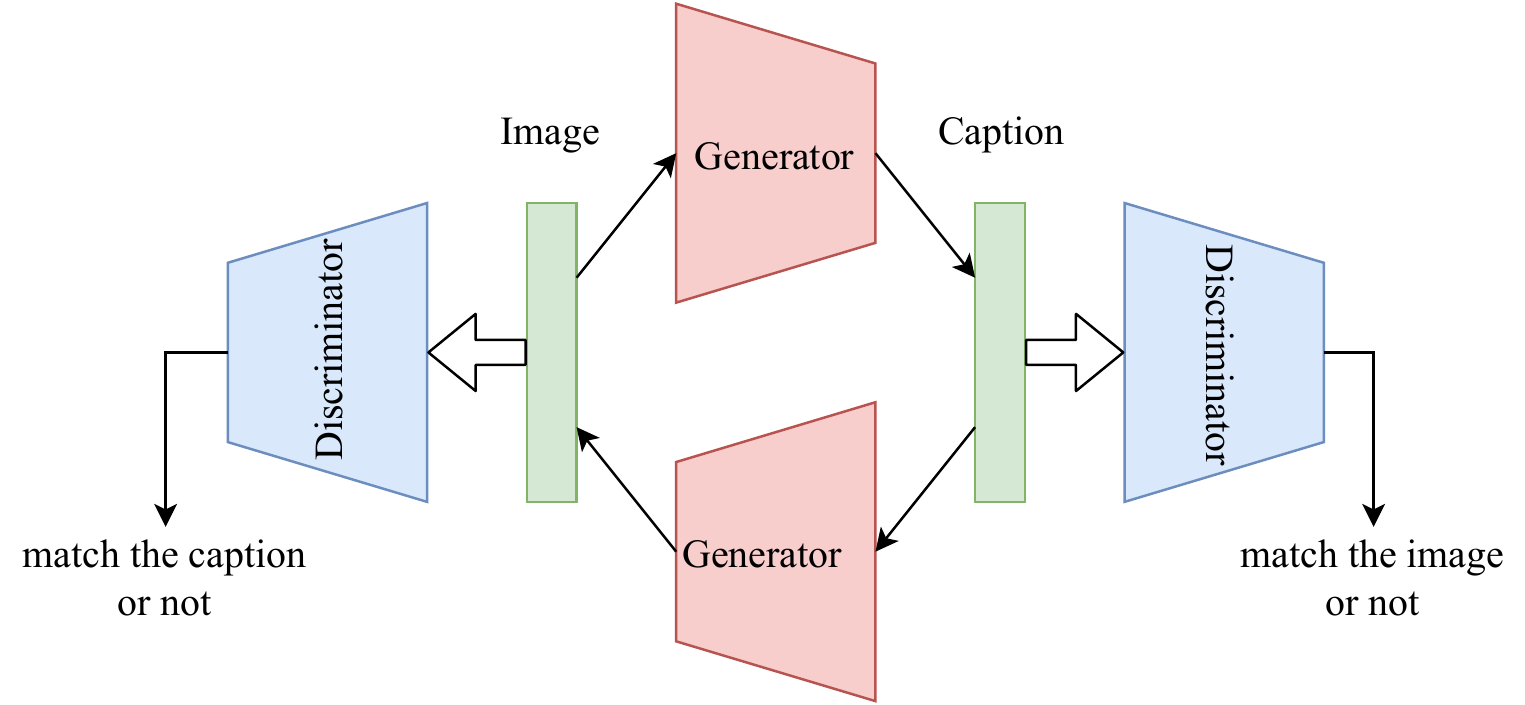}
 \caption{Simple diagram of the system of our proposed method.
}
 \label{fig:simple}
\end{figure}

A caption should faithfully describe an image. In this study, if the image recalled from a generated caption is close to the original image, then the caption is considered to represent the image faithfully. Thus, we propose a method to generate captions by learning to regenerate images using mutual transformations between images and texts. Previous research has been conducted to make it possible to search an original image from a generated sentence~\cite{luo2018discriminability}. However, in that case if only the minimum information necessary to describe the differences from other images for identification is available, then image retrieval can be considered possible. On the other hand, to generate an image from a caption generated from the input image, the caption must include information that is essential to convey the input image. In addition, the caption must make it possible to correctly understand the information when recalling an image from a caption. 
Therefore, an approach that attempts to generate captions from images by extracting the relationships between images and texts using a mutual transformation through machine learning can constitute a useful contribution. 

In this paper, we aimed to improve the performance of captions generated from images by incorporating a learning method using mutual transformations. A generated caption must be grammatically correct, such that it is easy for humans to read, and we assume that color information is not described for each pixel.
To realize such captions, we design a generator using generative adversarial networks (GANs). In addition, mutual transformations are performed between images and texts using a cycle-consistency loss. A simple diagram of the system of our model is presented in Figure~\ref{fig:simple}. There are two generators and two discriminators in this model. 

The contributions of this research are summarized as follows. 
\begin{itemize}
 \item We propose a method for image captioning through training to regenerate the input image. 
 \item The effectiveness of the proposed method is demonstrated by an automatic evaluation, and crowdsourcing to evaluate the degree to which images are described. 
\end{itemize}

\section{Related Work}
Here, we introduce related research on image-to-text generation, text-to-image generation, and mutual transformations between different domains using GANs. 
\subsection{Image Captioning}

Object recognition in images, understanding the relationships between objects, and the generation of grammatically correct sentences are required to generate captions from images, and research to make this possible has been conducted in recent years using machine learning. We refer the reader to the representative study of Vinyals et al.~\cite{vinyals2015show}. 
This represents an application of the encoder--decoder model, which achieves a strong performance in machine translation and uses a convolutional neural network (CNN) to encode feature from images and decode captions using a recurrent neural network (RNN). Here, long short-term memory (LSTM)~\cite{hochreiter1997long} is employed for the RNN. 
In addition, methods incorporating attention have recently become mainstream~\cite{xu2015show}. When generating a caption, the first word is acquired by performing discrete sampling, such as searching for the maximum value of the appearance probability of the next word generated from the image features. Then, the next word is acquired from that word in the same manner, and so on. 

In the method of Dai et al.~\cite{dai2017towards}, the authors combine the image captioning network of Vinyals et al. with a network for evaluating the generated captions. In the generator, the latent variable connected to an image feature extracted by the CNN is passed to LSTM. The generated caption and image are input into a network for evaluation, and the image features extracted by the CNN are multiplied by the text features extracted by LSTM to determine whether the caption matches the image. In the unsupervised image captioning method of Feng et al.~\cite{feng2018unsupervised}, the authors proposed an extension to unpaired data. 

\subsection{GANs for Text Generation}

GANs~\cite{goodfellow2014generative} consist of two parts: a generator and discriminator. 
In GANs, parameters of a generator are updated using gradients from a discriminator. However, because text generation usually involves indifferentiable discrete sampling, as in the study of Vinyals et al., it is not possible to propagate gradients from discriminators and train networks. 
With this motivation, SeqGAN was proposed by Yu et al.~\cite{yu2017seqgan} as a method enabling the generation of text from latent variables using GANs. As for ordinary GANs, the discriminator determines whether the input text is real or fake. An RNN is employed as the generator, but reinforcement learning is utilized for training. The parameter of the generator is utilized as the policy, the result of the discriminator acts as the reward, and the parameter is updated such that the reward is maximized using the policy gradient method. Because the discriminator can only calculate the reward for data series for which generation has been completed, the remainder of a partially generated series is generated using a Monte Carlo search, and the reward is calculated approximately. Using this method, it is difficult to start learning from random parameters, owing to the use of the policy gradient method, and pre-training is required. 

On the other hand, GANs for sequences of discrete elements with the Gumbel-softmax distribution (GSGAN), proposed by Kusner et al.~\cite{kusner2016gans}, utilizes the Gumbel-softmax~\cite{jang2016categorical} as a generator, which is differentiable. Here, the Gumbel-softmax is as shown in Equation~\ref{eq:gumbel-softmax}. $\pi_1, \pi_2, ..., \pi_k$ is the probability of each class before putting it in gumbel-softmax, $\tau$ is the temperature parameter. As $\tau$ is smaller, the output is closer to the one-hot vector, and as $\tau$ is larger, the output is closer to uniform distribution. 
Even if the text is generated in one go from the latent variable to the end, such as in the method of generating the first word from the latent variable and then generating each following word from the current word, parameter updating by error backpropagation is enabled.
Because the Gumbel-softmax can output the approximate one-hot vector, it can obtain the next predicted word by approximation of the one-hot expression without performing the process of searching for the maximum value. This achieves differential sampling, and trains a generator without using reinforcement learning. However, in this study only data sequences of short length, such as lists of symbols, are generated. An additional study employs the Gumbel-softmax to generate image descriptions using GANs~\cite{shetty2017speaking}. 
\begin{eqnarray}
  \label{eq:gumbel-softmax}
  \begin{aligned}
    u &\sim \rm{Uniform}(0,1) \\
    g &= -\log (-\log (u)) \\
    z_i &= \frac{\exp((\log(\pi_i) + g_i)/\tau)}{\sum_{j=0}^k \exp((\log(\pi_j) + g_j)/\tau)} \quad for \,\, i=1,...,k 
  \end{aligned}
\end{eqnarray}

\subsection{Text-to-Image Generation}

The method AlignDRAW~\cite{mansimov2015generating} was proposed by Mansimov et al. to generate images from text. This approach estimates the relationship between text features and a generated image using an RNN, but produces a blurred image that only captures rough features. 

Reed et al.~\cite{reed2016generative} generated images from text using GANs. The DCGAN~\cite{radford2015unsupervised} method, proposed by Radford et al., is a mainstream approach using GANs to generate images from latent variables, and employs a CNN. In addition, CGAN~\cite{mirza2014conditional} makes it possible to generate an image that matches a condition label. Using the same network as CGAN, in the method of Reed et al. the text feature vector extracted from a caption is connected to the latent variable as a condition, and input into the generator. The discriminator determines whether the generated image is valid for the caption. A method~\cite{reed2016learning} combining a CNN and RNN is also employed to extract features from captions. The model trained using this method is utilized as a text encoder, but this component is employed without updates when training GANs. 

The StackGAN method of Zhang et al.~\cite{zhang2017stackgan} is another approach for generating high-resolution images.
Like that of Reed et al., this method is based on CGAN, but the training is divided into two stages. An image satisfying the conditions is generated in the first stage, and a high-resolution image is generated in the second stage. Here, if the text feature $\varphi_t$ extracted from the caption $t$ is employed as it is, then the input vector is biased, and learning becomes unstable. To circumvent this issue, to maintain diversity the authors employ a condition $c$ sampled randomly from the normal distribution, whose average $\mu$ and diagonal covariance matrix $\Sigma$ are calculated from the text feature $\varphi_t$. In addition, to avoid overfitting the Kullback--Leibler divergence is incorporated into the loss function, representing the distance between the standard normal distribution and the normal distribution obtained from the feature. 
An additional study enables the generation of high-resolution images by end-to-end learning without division into two stages~\cite{zhang2018photographic}. 

\subsection{Transformation Using the Cycle Consistency}

The CycleGAN~\cite{zhu2017unpaired} method, proposed by Zhu et al., employs GANs for mutual transformations between images. The authors combined two GANs, and enabled color conversions of images such as horses and zebras, and summer and winter landscapes. In CycleGAN, let X and Y represent the image domains, such as summer and winter.
Then, we have four networks: the X to Y generator $G_Y$,
discriminator of Y $D_Y$,
Y to X generator $G_X$, and
discriminator of X $D_X$. 
In this model, the data of the domain Y is generated from that of the domain X, and the data of X is regenerated from the generated data. The data of X is generated from that of Y, and the data of Y is regenerated from the generated data in the same manner. 
CycleGAN utilizes two types of loss function for learning. The reason for employing two is that they combine two GANs. One is the same as for regular GANs. The other is called the cycle-consistency loss, and represents the error between the original data and the result obtained by passing this through the two generators. This is treated as a loss function for the generators, and is not used to train the discriminators. In this manner, CycleGAN enables mutual transformations of images, even for datasets that do not contain pair data. 
Text2image2textGAN, proposed by Gorti et al.~\cite{gorti2018text}, is an example of using the cycle-consistency to generate images from text. Here, an image is generated from a text, a text is generated from the resulting image, and learning is performed using the consistency loss between the regenerated and original texts. 
The model of Vinyals et al. is employed in the image-to-text generator, and so differentiable sampling as in GSGAN is not performed. 

\section{Proposed Model}

\begin{figure*}
 \centering
 \includegraphics[width=1.0\linewidth]{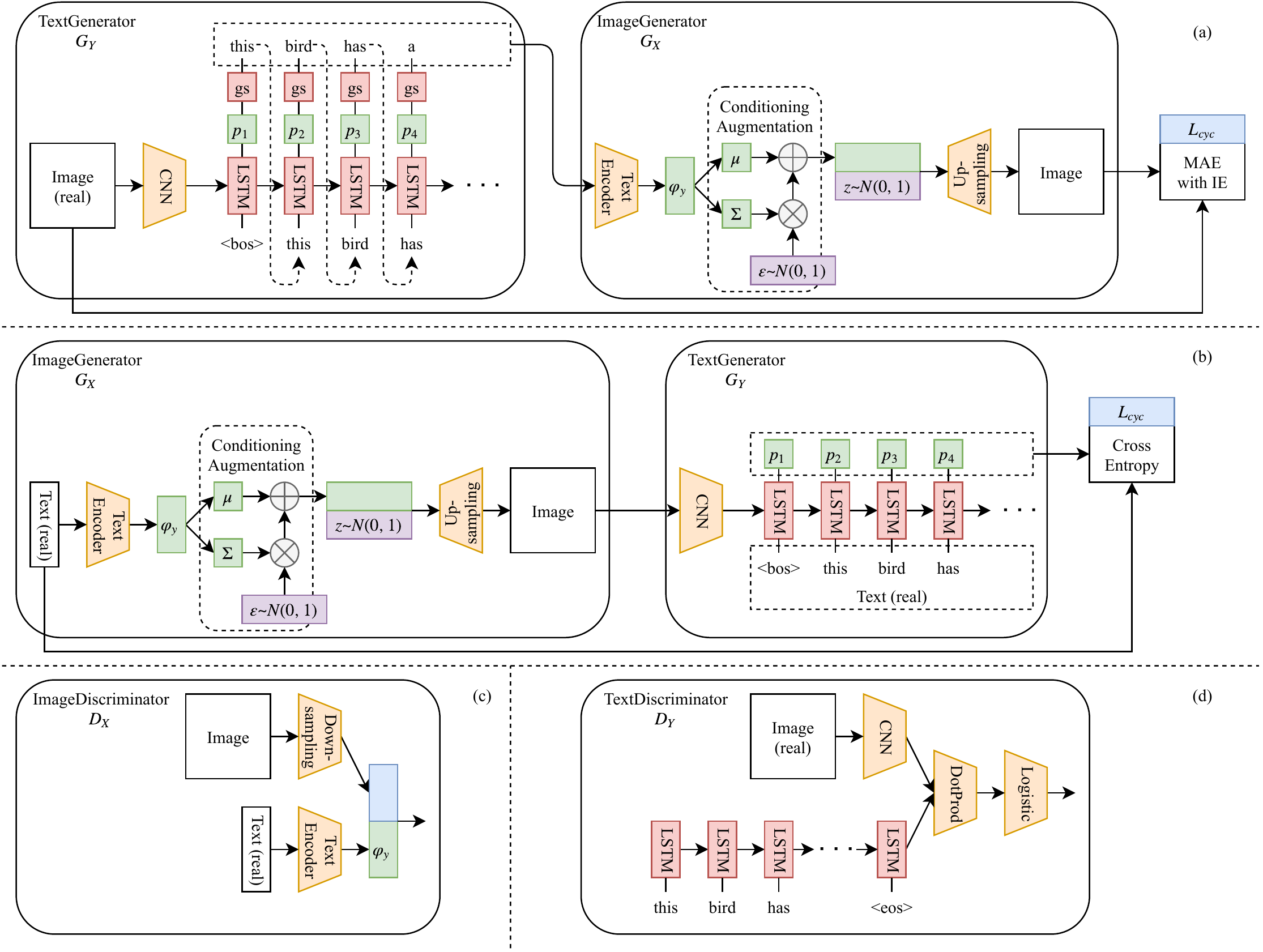}
 \caption{Our proposed model. The upper-left figure (a) illustrates the regeneration of an image. IE and gs in this figure indicate the image encoder and Gumbel--softmax. The lower-left figure (b) depicts the regeneration of a text. The upper-right figure (c) illustrates the discriminator for images, and the lower-right figure (d) illustrates the discriminator of texts. }
 \label{fig:allpipeline}
\end{figure*}

Our model performs mutual transformations between images and captions with reference to CycleGAN. The entire pipeline is shown in Figure~\ref{fig:allpipeline}. Unlike CycleGAN, which performs transformations between images, pre-training is performed to handle texts consisting of series data, so that a grammatically correct description and an image without noise can be generated. 
We explain each procedure in detail below. 

\subsection{Image-to-Caption GAN}

In GANs that generate captions from images, two types of generator are prepared. These generators share the same network parameters. 

One is employed to generate captions from images in the dataset. This generator is shown in Figure~\ref{fig:allpipeline} (a). Incorporating the method of GSGAN, we utilized the Gumbel-softmax to generate the text from an image feature. This is because a further image is generated from the generated caption, and error backpropagation can be performed during learning. 

The other generator is employed to generate a caption from a generated image. This generator is shown in Figure~\ref{fig:allpipeline} (b), and is based on the method of Vinyals et al. Here, the trained model of VGG16~\cite{simonyan2014very} was employed as a CNN for extracting features of images. An image is input into the trained model, and pooling is performed in the fifth layer. The output of this is then input into LSTM through the linear layer, and the features are extracted by the hidden vector of LSTM. 
The features are utilized as the first LSTM hidden vector of the caption generation component, the word representing the beginning of the sentence is employed as the first input, and the cross entropy of the probability distribution of the next word to be predicted and the ground truth are calculated. The first word of the ground truth is adopted as the next input, and cross entropy of the probability distribution of the next word to be predicted and the ground truth are again calculated. The generator is pre-trained by calculating the sum of the cross entropies until the end of the sentence in the same manner. 

A caption and an image are input into the discriminator. The captions generated by the generator are approximated as one-hot expressions by the Gumbel-softmax, and the sentences of the ground truth are represented by one-hot expressions. By multiplying the image features extracted by the trained VGG16 model with the caption features extracted by LSTM, it is determined whether captions match images. To prevent the discriminator from becoming overly strong, it is not pre-trained. 

When training both generators and classifiers, the parameters of the VGG16 component are not updated, as these are contained in the trained model. 

\subsection{Caption-to-Image GAN}

We designed the training of GANs that generate images from captions based on the first stage generator and discriminator of StackGAN. The loss function of the generator is the same as that of StackGAN, such that the amount of Kullback--Leibler information is added to the loss function of the normal generator of GANs. As with StackGAN, the discriminator also employs either the generated image or correct image from the dataset and the correct text features as input to determine whether an image matches the text. Here, the convolution and linear combination layers from the SNGAN~\cite{miyato2018spectral} method are adopted, such that spectral normalization is performed in the discriminator to stabilize the learning. 

Next, the resulting caption is input into a text encoder to extract features and generate an image from these features. In the method of Reed et al. and StackGAN, the text encoder employs a learned model, and parameters of the model are not updated while training for image generation. In this study, bi-directional LSTM~\cite{bahdanau2014neural} is employed as the text encoder, to extract the forward and backward features of the caption and connect them. While the effects of words increase at the end of the text in ordinary LSTM, the effects of words at the beginning of text can also be increased by employing bi-directional LSTM. This text encoder was trained by the generator using its loss function. 

The generator, discriminator, and text encoder were pre-trained. 

\subsection{Image-to-Caption Mutual Transformation}

For the mutual transformation, it is necessary to calculate the cycle-consistency loss from the results through the two generators and the original input, and to utilize the loss for learning. 
We used two types of L1 loss, for the cycle-consistency loss between images (Figure~\ref{fig:allpipeline} (a)) and the cross entropy for the cycle-consistency loss between captions (Figure~\ref{fig:allpipeline} (b)).

Images should not only be directly compared on a pixel-by-pixel basis, but the contents of the images should also be compared. Therefore, the L1 loss calculated from a direct image and that calculated from the value of the intermediate layer passing the image through VGG16 were incorporated. 
Specifically, we employed the calculation results from up to the fifth pooling layer. In addition, weighting was performed using hyper parameters, so that the effect of the L1 loss calculated through VGG16 was greater than that of the directly calculated L1 loss. The technique of comparing images through the image encoder and calculating the loss is also employed in SRGAN~\cite{ledig2017photo}, making it possible to prevent blurry outputs, such as those occurring when averaging pixel values when comparing images directly. 

For the cycle-consistency loss between captions, we did not employ the generated text as it was, but rather we used the vector before passing the Gumbel-softmax, which represented the probability distribution of the predicted word, and calculated the cross entropy using the captions of the dataset. 

\subsection{Objective Function}

Equation~\ref{eq:cycle_text} defines the objective function of GANs that generates captions from images. Here, the domain X is an image and Y is a text. Furthermore, $G_Y$ and $D_Y$ are the generator and discriminator of captions, respectively. 
\begin{eqnarray}
  \label{eq:cycle_text}
  \begin{split}
  \max_{D_Y} L_{D_Y} &= E_{({\bm x}, {\bm y}) \sim p_{data}({\bm x}, {\bm y})}[\log D_Y({\bm y}, {\bm x})] \\ &+ E_{{\bm x} \sim p_{data}({\bm x})}[\log (1-D_Y(G_Y({\bm x}), {\bm x}))] \\
  \min_{G_Y} L_{G_Y} &= E_{{\bm x} \sim p_{data}({\bm x})}\left[-\log \frac{D_Y(G_Y({\bm x}))}{1-D_Y(G_Y({\bm x}))}\right]
  \end{split}
\end{eqnarray}

Equation~\ref{eq:cycle_image} defines the objective function of GANs that generates images from captions. Here, $G_X$ and $D_X$ are the generator and discriminator of images, respectively. Furthermore, $\varphi_{\bm y}$ denotes features extracted by the text encoder. The text encoder component is trained using only the generator's loss function. Here, the hyper parameter was set as $\lambda_{KL}=2.0$. 

\begin{eqnarray}
  \label{eq:cycle_image}
  \begin{split}
  \max_{D_X} &L_{D_X} = \\ E&_{({\bm x}, {\bm y}) \sim p_{data}({\bm x}, {\bm y})}[\log D_X({\bm x}, \varphi_{\bm y})] + \\ E&_{{\bm z} \sim p_{z},{\bm y} \sim p_{data}({\bm y})}[\log (1-D_X(G_X({\bm z}, \varphi_{\bm y}), \varphi_{\bm y}))] \\
  \min_{G_X} &L_{G_X} = \\ E&_{{\bm z} \sim p_{z},{\bm y} \sim p_{data}({\bm y})}[\log (1-D_X(G_X({\bm z}, \varphi_{\bm y}), \varphi_{\bm y}))] + \\ \lambda&_{KL} D_{KL} (N(\mu(\varphi_{\bm y}),\Sigma(\varphi_{\bm y})) \| N(0,1))
  \end{split}
\end{eqnarray}

Finally, Equation~\ref{eq:cycle_cons} defines the objective function of the cycle-consistency loss. Here, $F_{IE}$ is the image encoder, and we employed VGG16. Furthermore, ${\bm y}_t$ is the $t$-th word in a caption and $T$ is the length of a caption. 
Here, the hyper parameters were set as $\lambda_1=1.0$, $\lambda_2=1,000$, and $\lambda_3=0.01$. 
\begin{eqnarray}
  \label{eq:cycle_cons}
  \begin{split}
  &\min_{G_Y,G_X} L_{cyc} \\ &= E_{{\bm x} \sim p_{data}({\bm x})}[\lambda_1 \| G_X(G_Y({\bm x})) - {\bm x} \|_1] \\ &+ E_{{\bm x} \sim p_{data}({\bm x})}[\lambda_2 \| F_{IE}(G_X(G_Y({\bm x}))) - F_{IE}({\bm x}) \|_1] \\ &+ E_{{\bm y} \sim p_{data}({\bm y})}[-\lambda_3 \sum_{t=0}^T {\bm y}_t\log (G_Y(G_X({\bm y}_t)))] 
  \end{split}
\end{eqnarray}

In summary, the objective function of the proposed method is defined as in Equation~\ref{eq:cycle_all_main}. 
\begin{eqnarray}
  \label{eq:cycle_all_main}
  \begin{split}
    \max_{D_Y,D_X} V_{D_Y,D_X} &= L_{D_Y} + L_{D_X} \\
    \min_{G_Y,G_X} V_{G_Y,G_X} &= L_{G_Y} + L_{G_X} + L_{cyc}
  \end{split}
\end{eqnarray}

\section{Experiments}

Here, we describe an experiment performed to generate a caption from an image, and the generated caption is evaluated. 

\subsection{Datasets}

\begin{figure*}
 \centering
 \includegraphics[width=1.0\textwidth]{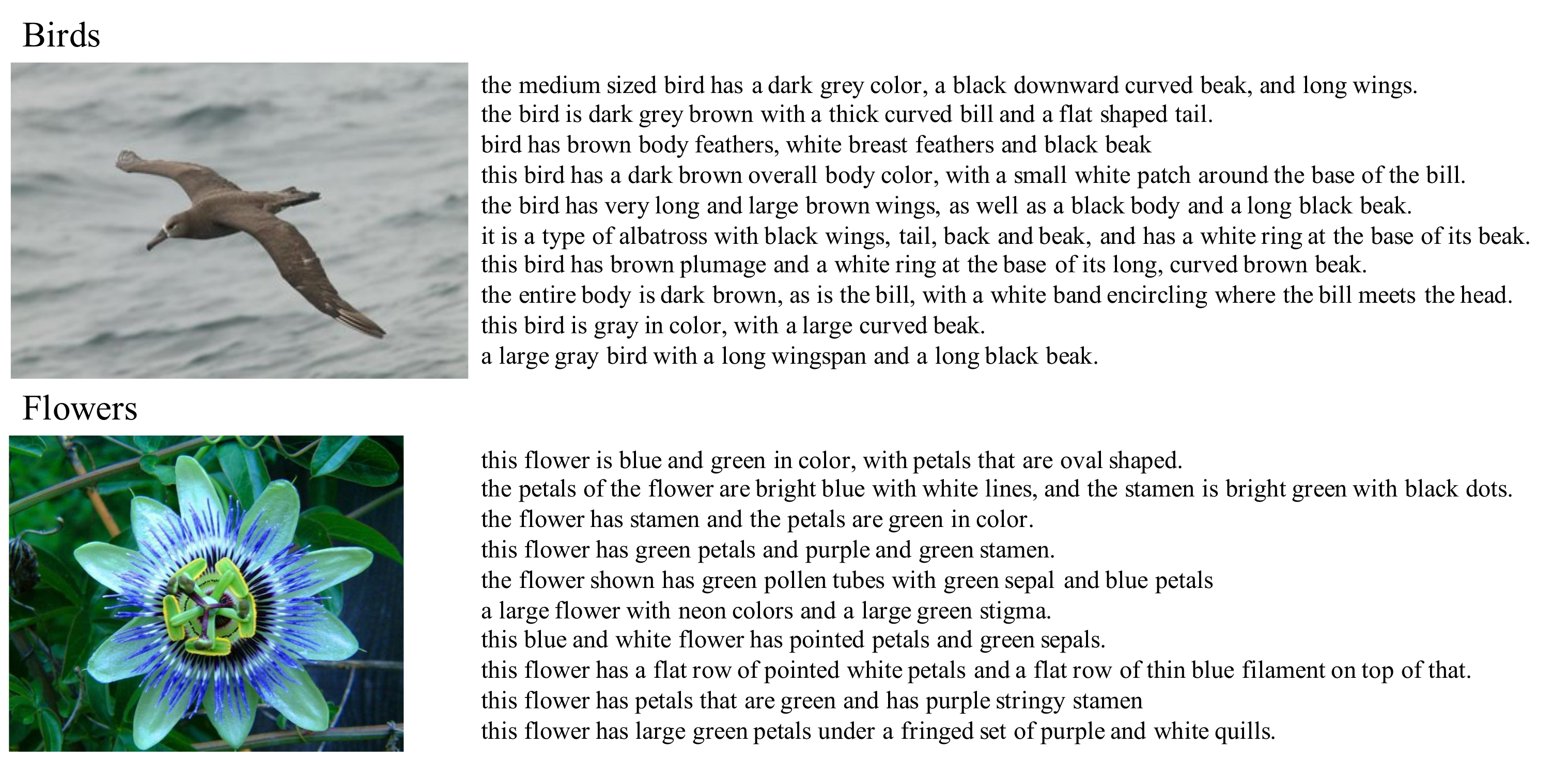}
 \caption{Some examples of the dataset provided by Reed et al.}
 \label{fig:dataset}
\end{figure*}

We employed a dataset combining Caltech-UCSD Bird~\cite{wah2011caltech}, consisting of bird images, and Oxford-102 Flower~\cite{nilsback2006visual}, consisting of flower images. The dataset contains attached captions, and was provided by Reed et al.~\cite{reed2016learning}. Unlike in CycleGAN, this is a paired dataset. Because this dataset is limited to birds and flowers, it is often employed to study image generation from text. Therefore, we considered it to be appropriate for this study. 
Some examples are presented in Figure~\ref{fig:dataset}. 

Here, 10 captions were attached to each image using Amazon Mechanical Turk (AMT)~\cite{mturk}. Each of these 10 sentences is independent. Workers ware instructed not to mention background, so the information that said it was in the sea or forest was not included in the captions. 

There are a total of 11,788 images of birds in 200 classes, and a total of 8,188 images of flowers in 102 classes. Of these, 90\% were utilized as training data and 10\% as test data. Specifically, 10,609 training and 1,179 test data items were from the bird dataset, and 7,369 training and 819 test data items were from the flower dataset. 

\begin{figure}[t]
 \begin{center}
  \includegraphics[width=.5\textwidth]{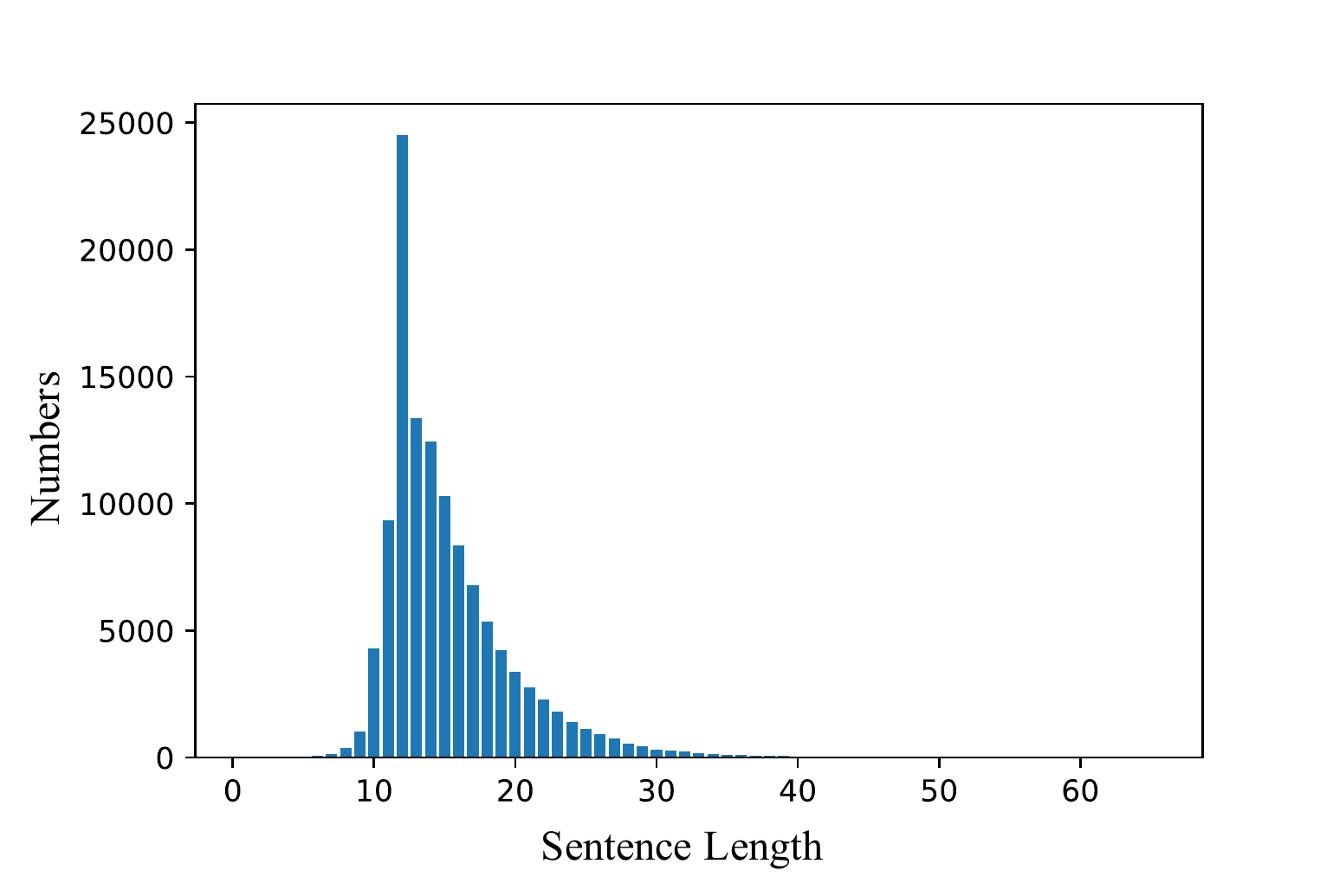}\\
  \vspace*{5pt}
  \includegraphics[width=.5\textwidth]{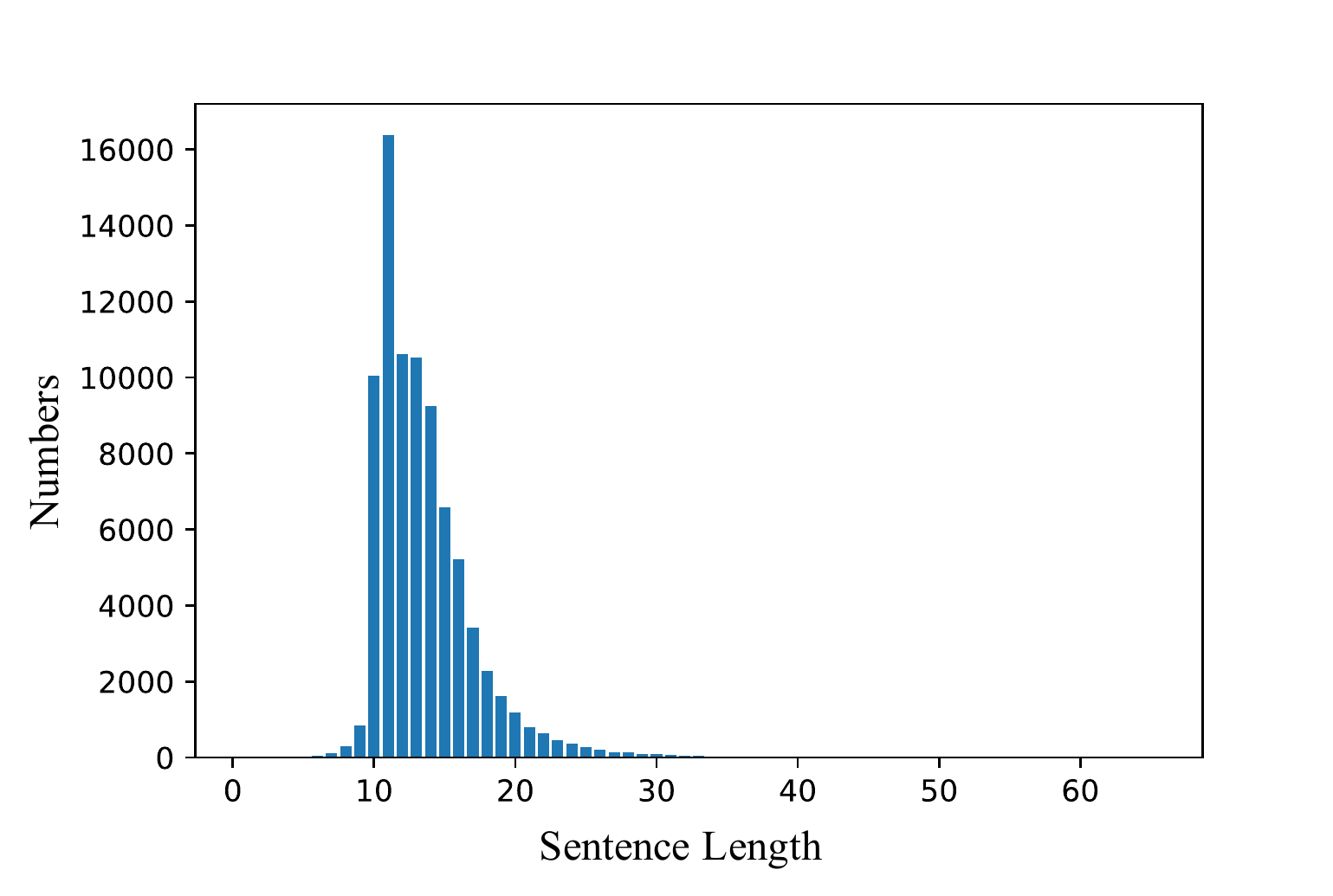}
  \caption{Lengths of captions in the dataset. The upper and lower parts represent the bird and flower data, respectively. }
  \label{fig:length}
 \end{center}
\end{figure}

Here, the lengths of the captions included in each dataset are shown in Figure~\ref{fig:length}. The bird and flower datasets vary in length. 

Therefore, to make the lengths of the utilized captions uniform, one of the 10 captions for each image was selected to be utilized according to the following method.
\begin{itemize}
 \item Remove periods, commas, and semicolons from the captions.
 \item Align all captions to 20 words in length by filling the end of the caption with a symbol indicating the end of the sentence or deleting the 21st and later words.
 \item During training, select one of the 10 captions at random for each iterator (however, the caption shall be limited to five or more words except for the sentence-ending symbol).
\end{itemize}
The resulting captions of the dataset were utilized in the experiment. In addition, because the bird dataset was provided with bounding-box information indicating the part of the image where the bird appears, this part was cropped and utilized. 
The images were RGB representations of size $64 \times 64$. 

\subsection{Experimental Setup}
We generated captions of 20 words in length from $64 \times 64$-sized images. 
When a caption was input into the text encoder, consisting of bi-directional LSTM, the text features were extracted as a 1,024-dimensional vector, in which forward vector and backward vectors of 512 dimensions were connected, and size-$64 \times 64$ image was generated from the text features and a 100-dimensional latent variable according to the normal distribution. 
The training consisted of 500 epochs in pre-training and 200 in subsequent training. Adam~\cite{kingma2014adam} and weight decay were employed for optimization during the whole training period, including pre-training. 

As a comparison method, image captioning and image regeneration were performed by combining $G_X$ and $G_Y$, which were trained the same number of times as the proposed method, without using the cycle-consistency loss. 

\subsection{Qualitative Results}

\begin{figure*}
 \centering
 \includegraphics[width=0.75\textwidth]{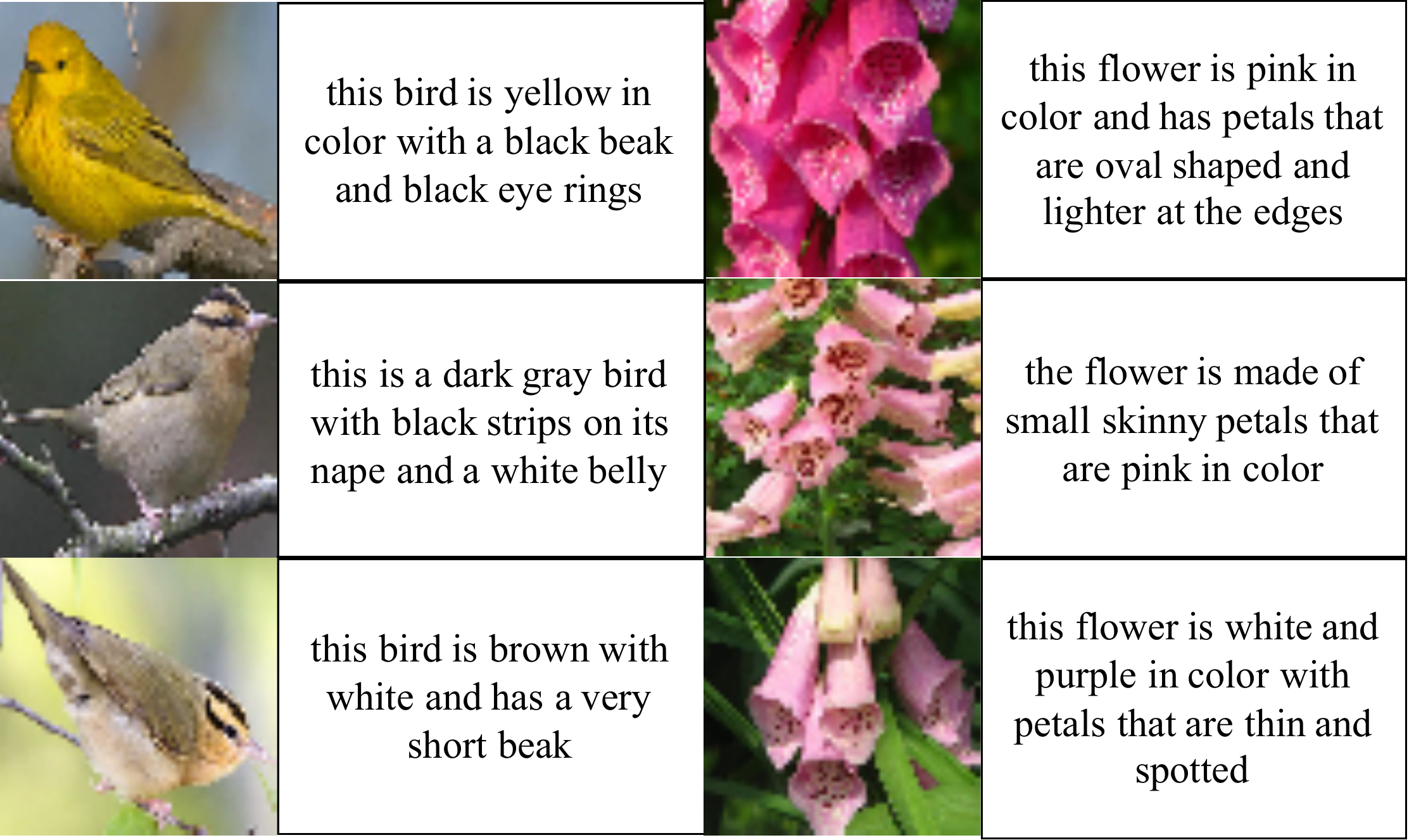}
 \caption{Examples of successes. Each caption on the right is generated from the image on the left. }
 \label{fig:success}
\end{figure*}

Some successful examples from this experiment are presented in Figure~\ref{fig:success}. The captions were generated using the proposed method from images in the dataset for test. When the symbol indicating the end of the sentence was output, words after that were deleted. They correctly describe the colors of birds or flowers overall, and also mention the colors of beaks or shapes of petals. Furthermore, the captions are also grammatically correct, at a level that can be understood by humans.

\subsection{Caption Evaluation}

As a quantitative evaluation method for the captions generated from the test data, automatic metrics such as BLEU~\cite{papineni2002bleu}, ROUGE~\cite{lin2004rouge}, Meteor~\cite{denkowski2014meteor}, and CIDEr~\cite{vedantam2015cider} are employed. 
The results are shown in Table~\ref{table:BLEU}. For both the bird and flower datasets, the proposed method achieved a comparable or better performance than the comparison method. In particular, the score improvement in BLEU-4 and CIDEr was remarkable. 

\begin{table}[t]
  \caption{Automatic evaluation of generated captions}
  \label{table:BLEU}
  \begin{center}
    \small
    \begin{tabular}{c|c|c|c|c} \hline
       & BLEU-4 & ROUGE & Meteor & CIDEr \\ \hline \hline
      Ours(bird) & $\bm{0.057}$ & $\bm{0.304}$ & 0.146 & $\bm{0.218}$ \\ \hline
      Comp.(bird) & 0.047 & $\bm{0.304}$ & $\bm{0.147}$ & 0.207 \\ \hline \hline
      Ours(flower) & $\bm{0.070}$ & $\bm{0.294}$ & $\bm{0.146}$ & $\bm{0.212}$ \\ \hline
      Comp.(flower) & 0.062 & 0.287 & 0.141 & 0.190 \\ \hline
    \end{tabular}
  \end{center}
\end{table}

\subsection{Human Evaluation}

To determine whether faithful captions were generated, a manual evaluation was also performed by crowdsourcing using AMT. Given generated caption pairs (from the proposed and comparison methods) together with an image from the dataset, workers were asked to select which caption better represents the image. Here, we instructed the workers to give priority to judging the extent to which the caption refers to the image, rather than focusing on grammatical order. The data employed here consisted of 100 randomly selected items from the test data, and five workers completed the same task. The workers were all certified as Master by AMT. The points obtained by summing the selections of each option for the proposed and comparison methods were used for the evaluation. In addition, binomial tests were performed on the evaluation results, to determine whether the differences between the proposed and comparison methods were significant. Here, it was tested whether the proportion selecting the proposed method was significantly different from 0.5, with a significance level of 0.05. The results are shown in Table~\ref{table:AMT}. For both datasets, the proposed method achieved higher points, and the statistical tests show a significant difference between the proposed and comparison methods. (Each p value was less than 0.05.) 

\begin{table}[t]
  \caption{Evaluation of faithfulness using crowdsourcing. Workers chose our methods more than comparison method. }
  \label{table:AMT}
  \begin{center}
    \begin{tabular}{c|c} \hline
       & faithfulness of generated captions \\ \hline \hline
      Ours(bird) & $\bm{318}$ \\ \hline
      Comp.(bird) & 182 \\ \hline
      P value & 1.247e-09 \\ \hline \hline
      Ours(flower) & $\bm{310}$ \\ \hline
      Comp.(flower) & 190 \\ \hline
      P value & 8.963e-08 \\ \hline
    \end{tabular}
  \end{center}
\end{table}

\subsection{Regenerated Image Evaluation}

As a quantitative evaluation method for regenerated images, the fine-tuned inception model~\cite{zhang2017stackgan} of StackGAN was employed, and the inception score was calculated. Specifically, we employed the fine-tuned inception model of Zhang et al.~\cite{szegedy2016rethinking}
for Caltech-UCSD Bird and Oxford-102 flower. The results are shown in Table~\ref{table:Inception}. The overall results are not considerably different, and for the flower dataset the score was lower for the proposed method than for the comparison method. 

\begin{table}[t]
  \caption{Inception scores of regenerated images}
  \label{table:Inception}
  \begin{center}
    \small
    \begin{tabular}{c|c|c} \hline
       & mean & std. \\ \hline \hline
      Ours(bird) & $\bm{1.46}$ & 0.14 \\ \hline
      Comp.(bird) & 1.38 & 0.10 \\ \hline \hline
      Ours(flower) & 1.26 & 0.07 \\ \hline
      Comp.(flower) & $\bm{1.34}$ & 0.08 \\ \hline
    \end{tabular}
  \end{center}
\end{table}

\subsection{Extension to unpaired data}
The dataset used in this experiment is used as a non-pair data set by randomly selecting image and description pairs.
Learning was also performed using non-paired data sets in comparison methods and pre-training. The experiment was performed with the dataset only being unpaired in exactly the same experimental setting, and the results are shown in Table~\ref{table:unpair}. Although there are also cases where the proposed method outperforms the comparison method, there was no significant difference. The results in this experiment are not significantly inferior in performance compared to when the experiment was performed in pair dataset. However, the dataset used here is the one in which the domain is limited to birds or flowers, so we do not know about the case where we experimented with a dataset containing more versatile images. As a matter of discussion, even if we create a non-pair dataset at random like this time, it is possible that the selected captions partially become similar with correct sentences, and there is almost no difference. 

\begin{table}[t]
  \caption{Automatic evaluation of generated captions(unpaired)}
  \label{table:unpair}
  \begin{center}
    \small
    \begin{tabular}{c|c|c|c|c} \hline
       & BLEU-4 & ROUGE & Meteor & CIDEr \\ \hline \hline
      Ours(bird) & 0.056 & 0.308 & 0.149 & 0.216 \\ \hline
      Comp.(bird) & 0.048 & 0.299 & 0.144 & 0.197 \\ \hline \hline
      Ours(flower) & 0.065 & 0.295 & 0.142 & 0.185 \\ \hline
      Comp.(flower) & 0.064 & 0.282 & 0.145 & 0.181 \\ \hline
    \end{tabular}
  \end{center}
\end{table}

\section{Conclusions}

In this study, we proposed a learning method for image captioning using mutual transformations between images and texts. We demonstrated that mutual transformations are possible using the cycle-consistency loss. Furthermore, automatic evaluations demonstrated that a comparable or better performance was achieved by the proposed method with paired dataset. In addition, crowdsourcing evaluations also demonstrated that our method is more effective than one without the cycle-consistency loss. 

In order to improve performance more, it is necessary to increase the accuracy of GANs used for mutual conversion. In particular, GANs that generate images matching text from text are considered to be difficult to achieve higher accuracy than GANs that generate text from images in terms of the amount of information. Also, in this paper, we did not consider where to pay attention to images and captions. During learning, two types of loss function can interfere with parameter updates of each other. In order to avoid that, it is effective to find something better than the one used in this paper as a cycle-consistency loss between images. 

In order to know the effect on non-paired data sets, it is considered that datasets with more domains to handle are required. 

\section{Acknowledgement}
This work was partially supported by JST CREST Grant Number JPMJCR1403, Japan. 
We would like to thank Mikihiro Tanaka, Atsuhiro Noguchi, Hiroaki Yamane for helpful discussions. 

{\small
\bibliographystyle{ieee}
\bibliography{egbib}
}

\end{document}